%% file: main_for_arxiv.tex
\documentclass[lettersize,journal]{IEEEtran}
\usepackage{amsmath,amsfonts}
\usepackage{algorithmic}
\usepackage{array}
\usepackage{textcomp}
\usepackage{stfloats}
\usepackage{url}
\usepackage{verbatim}
\usepackage{graphicx}
\def\BibTeX{{\rm B\kern-.05em{\sc i\kern-.025em b}\kern-.08em
    T\kern-.1667em\lower.7ex\hbox{E}\kern-.125emX}}
\usepackage{balance}

\usepackage{xcolor}
\definecolor{iccvblue}{rgb}{0.21,0.49,0.74}
\usepackage[pagebackref,breaklinks,colorlinks,allcolors=iccvblue]{hyperref}

\usepackage{xspace}
\usepackage{multirow}  
\usepackage[utf8]{inputenc} 
\usepackage[T1]{fontenc}    
\usepackage{url}            
\usepackage{booktabs}       
\usepackage{amsfonts}       
\usepackage{nicefrac}       
\usepackage{microtype}      
\usepackage{colortbl}
\usepackage{graphicx}
\usepackage{pifont}
\usepackage{subcaption}
\usepackage{cleveref}
\usepackage{tcolorbox}

\usepackage{subcaption}

\crefname{figure}{Fig.}{Figs.}
\Crefname{figure}{Fig.}{Figs.}
\crefname{table}{Tab.}{Tabs.}
\Crefname{table}{Tab.}{Tabs.}

\definecolor{demphcolor}{RGB}{90,90,90}

\newcommand{\tablestyle}[2]{\setlength{\tabcolsep}{#1}\renewcommand{\arraystretch}{#2}\centering\footnotesize}

\newlength\savewidth

\newcommand{\cxmark}{\ding{55}}
\newcommand{\methodname}{MVP-Integrator\xspace}

\begin{document}
\title{MAC: A Benchmark for Multiple Attributes Compositional Zero-Shot Learning
\thanks{This work has been submitted to the IEEE for possible publication. Copyright may be transferred without notice, after which this version may no longer be accessible.}
}

\author{
    Shuo Xu,
    Sai Wang,
    Xinyue Hu,
    Yutian Lin,
    Sibei Yang,
    Yu Wu,
    Bo Du%
    \thanks{Shuo Xu, Sai Wang, Xinyue Hu, Yutian Lin, Yu Wu, and Bo Du are with the School of Computer Science, Wuhan University, Wuhan, China. (e-mail: shuoxu@whu.edu.cn; saiwang23@whu.edu.cn; wuyucs@whu.edu.cn; bodu@whu.edu.cn; yutian.lin@whu.edu.cn)} \thanks{Sibei Yang is with the School of Information Science and Technology, ShanghaiTech University, Shanghai, China.}%
    \thanks{Corresponding authors: Yutian Lin, Bo Du}
}

\maketitle

\begin{abstract}
Compositional Zero-Shot Learning (CZSL) aims to learn semantic primitives (attributes and objects) from seen compositions and recognize unseen attribute-object compositions. Existing CZSL datasets focus on single attributes, neglecting the fact that objects naturally exhibit multiple interrelated attributes. 
Their narrow attribute scope and single attribute labeling introduce annotation biases, misleading the learning of attribute and causing inaccurate evaluation.
To address these issues, we introduce the Multi-Attribute Composition (MAC) dataset, encompassing 22,838 images and 17,627 compositions with comprehensive  attribute annotations. 
MAC shows a complex relationship between attributes and objects, with each attribute type linked to an average of 82.2 object classes, and each object type associated with 31.4 attribute classes.
Based on MAC, we propose multi-attribute compositional zero-shot learning that requires deeper semantic understanding and advanced attribute associations, establishing a more realistic and challenging benchmark for CZSL. 
We propose Multi-attribute Visual-Primitive Integrator (\methodname), a robust baseline for multi-attribute CZSL, which disentangles semantic primitives and performs effective visual-primitive association. Experiments demonstrate that \methodname significantly outperforms existing CZSL methods on MAC with improved  efficiency. The dataset will be released publicly upon publication. The code and data are available at: \href{https://github.com/xs1317/MAC}{https://github.com/xs1317/MAC}.
\end{abstract}


\section{Introduction}
Compositional zero-shot learning (CZSL)~\cite{anwaar2022leveraging,  mancini2021open, atzmon2020causal, jiang2025compact} aims to compose knowledge of learned semantic primitives (e.g., attributes and objects) to recognize unseen compositions during the testing phase, where each attribute and object has been encountered during training. 
It highlights the model’s capacity to generalize knowledge and infer novel concepts from known primitives, mimicking human cognitive abilities. Additionally, it serves as the foundation for various downstream tasks such as open-vocabulary object recognition~\cite{bravo2023open, wang2023learning} and human object interaction~\cite{baek2025vhoip, wang2020learning}.

\begin{figure}[t]
    \centering    
    \begin{subfigure}[t]{1.0\columnwidth}
        \centering
        \includegraphics[width=\linewidth]{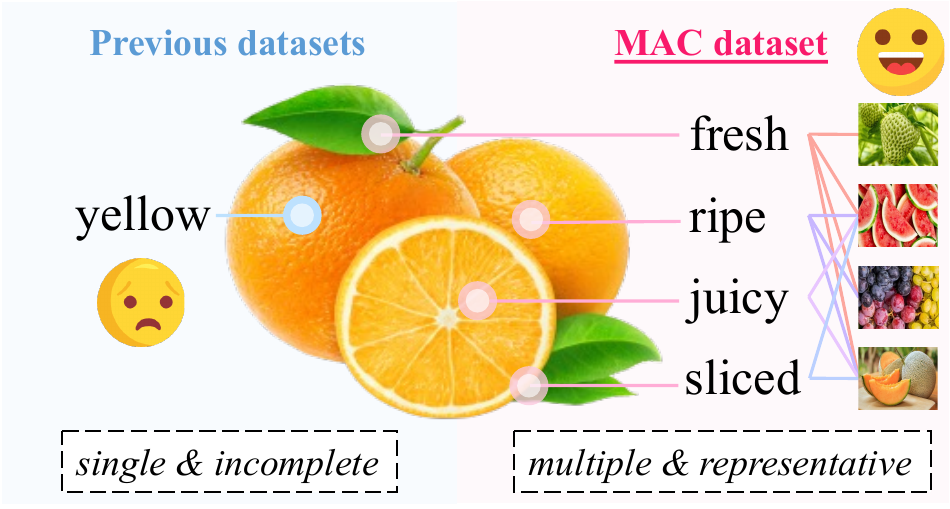}
        \vspace{-5mm}
        \caption{Comparison to previous datasets.}
        \label{fig:intro_1a}
    \end{subfigure}
    
    \vspace{2mm}
    
    \begin{subfigure}[t]{1.0\columnwidth}
        \centering
        \includegraphics[width=\linewidth]{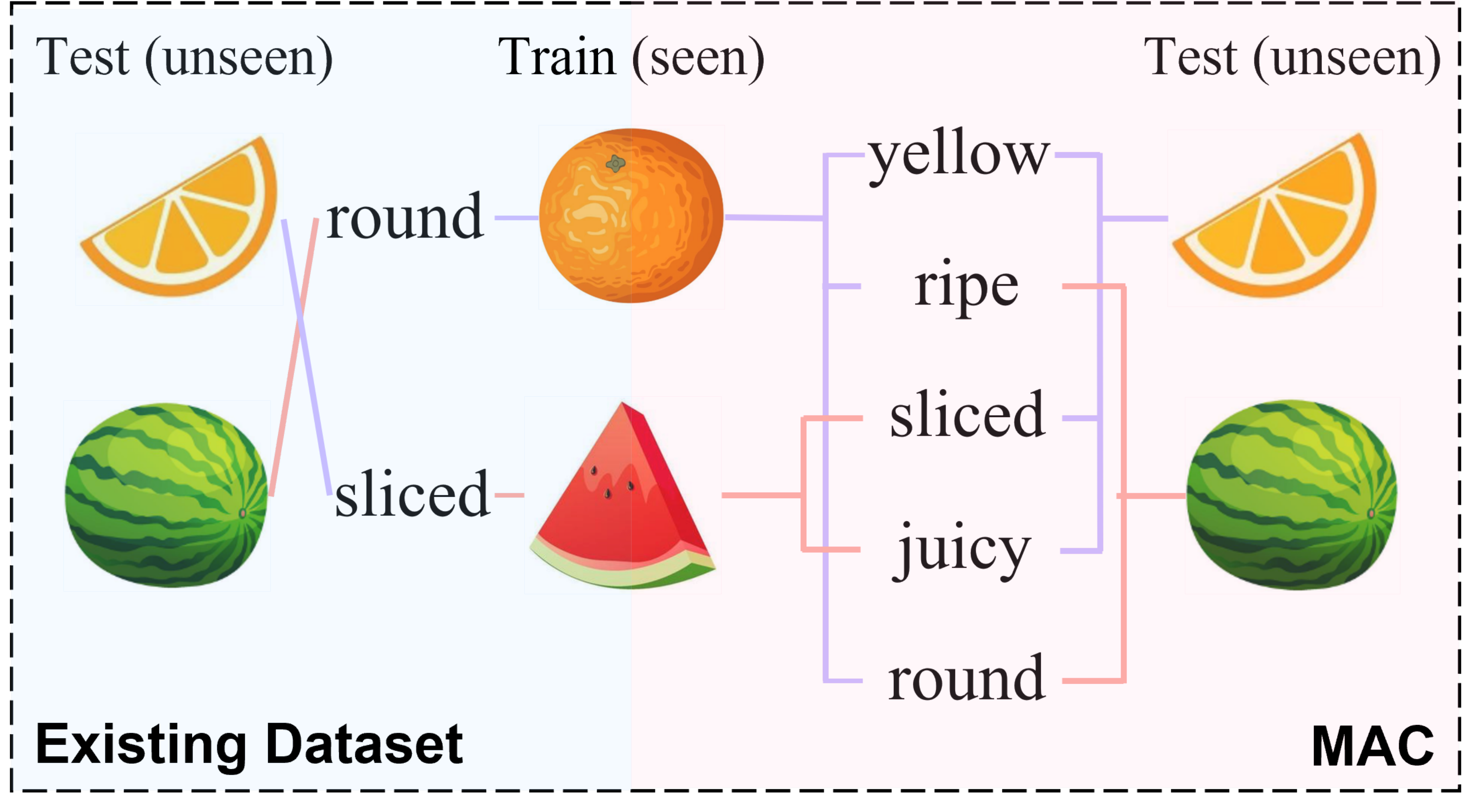}
        \vspace{-5mm}
        \caption{Task description.}
        \label{fig:intro_1b}
    \end{subfigure}
    \caption{(a) Compared to previous datasets. MAC provides comprehensive and representative attributes for the objects. (b) Multiple attribute compositional zero-shot learning. }
    \label{fig:intro_1}
    \vspace{-7mm}
\end{figure}

Existing CZSL datasets, such as MIT-States~\cite{isola2015discovering}, UT-Zappos~\cite{yu2014fine}, and C-GQA~\cite{naeem2021learning}, provide single-attribute annotations, neglecting the fact that real-world objects typically exhibit multiple interrelated attributes. 
Single-attribute annotation significantly hinders the model's understanding of the object's attributes, as single-label classification losses often suppress correct but unlabeled attributes.
Single-attribute annotation also leads to inaccurate evaluation of the model performance.
It not only causes valid predictions to be misjudged as incorrect but also hinders a comprehensive evaluation of model performance. 
Even when a prediction does not match the single-attribute ground truth, it may still be valid. For example, the object in \Cref{fig:intro_1a} can be described as ``yellow orange'' based on its color or as ``fresh juicy orange'' based on its state.
Besides, these datasets neglect the co-occurrence relationship of attributes. For instance, ``cool'' and ``refreshing'' frequently appear together in images of ice-cold cola. Introducing such correlations could enhance model representation learning.
Additionally, the images of these datasets for the same object class often demonstrate similar appearance features resulting in low complexity in composition space as shown in \Cref{tab:dataset}.

\input{tabs/tab_dataset.tex}

To address these limitations, we design a new dataset:  Multi-Attribute Composition (MAC) dataset. MAC includes 22,838 images and 17,627 compositions, with 60.76\% of the images annotated with three or more attributes. Compared to existing datasets, MAC is better suited for CZSL due to following characteristics.
\textbf{(1) Comprehensive Annotation.} 
Comprehensive attribute annotation helps the model learn attribute more effectively and reduces misjudgments during evaluation.
However, identifying all suitable attributes from a given set is challenging.
We initialize three attribute candidate sets (search keywords, CLIP, and GPT-4o) and ensure completeness and accuracy through manual checking and manual supplementation.
\textbf{(2) Rich Image Diversity.} 
Existing datasets primarily feature common attribute-object compositions, neglecting sparse relationships. 
For example, while concepts like ``blue strawberry'' are scarcely observed in real-world scenarios, such unusual compositions can  provide hard samples for evaluating compositional reasoning capabilities.
Current datasets also exhibit visual homogeneity within each object class, limiting models' ability to learn robust attribute associations. We further analyze this phenomenon in section C of the Appendix.
MAC utilizes LLM to generate diverse compositions for each object type, then crawls images from search engines to ensure visual variety.
\textbf{(3) Representative and Generalizable Attributes.} 
The attribute set directly determines the scope of the dataset.
Attributes must be representative to effectively describe an object's appearance. Additionally, they should generalize across diverse objects to  prevent models from inferring attributes solely based on object classes.
MAC leverages LLM knowledge to carefully select 99 representative and generalizable attributes, spanning low-level patterns (e.g., ``grainy'') to functional states (e.g., ``folded''). As in \Cref{tab:dataset}, each attribute type in MAC can be applied to 82.18 object classes on average, showing complex attribute-object relationship, significantly outperforming existing datasets.


We propose multiple attribute CZSL to better evaluate the model's compositional understanding. We use six metrics to evaluate the accuracy and completeness of models in  closed-world and open-world settings.
For consistency with existing works,  we also report results on AUC, Seen, and Unseen following \cite{naeem2021learning}. Nine methods are implemented on MAC, covering graph models~\cite{naeem2021learning}, prompt learning~\cite{zhou2022learning,nayak2022learning}, semantic disentanglement~\cite{huang2023troika,lu2023decomposed} and VLM~\cite{liu2024llava}. 

Existing CLIP-based methods are suboptimal for multiple attribute CZSL especially in the open-world setting. 
Their reliance on composition branch leads to a significant performance drop in the open-world setting. 
The composition branch also increases the inference complexity to $|A|\times|O|$ as it needs to calculate all the text representations for each composition. Finally, these methods often overlook the inherent semantic relationships between attributes, such as the co-occurrence of "shiny" and "clean". To solve these problems, we propose the Multi-attribute Visual-Primitive Integrator (\methodname), which eliminates  composition branch and leverages dual-branch prompt tuning to learn primitives. MVP-Integrator processes the image representations and text representations of primitives jointly. It uses self-attention~\cite{vaswani2017attention} on concatenated image-text representations to effectively capture attribute-object, attribute-attribute, and image-text relationships. It overcomes the relationship modeling challenges faced by previous methods in the large composition space. Due to the elimination of composition branch, \methodname inherently achieves consistent performance on open-world  and closed-world settings and significantly reduces the inference complexity to $|A|+|O|$. Extensive experiments show that \methodname outperforms existing methods with improved efficiency. 
In summary,  our main contributions are threefold:

\begin{itemize}
\item We introduce Multi-Attribute CZSL, a new task that extends CZSL to more realistic multi-label scenarios.

\item We construct the MAC dataset, featuring comprehensive attribute annotations and diverse visual content, addressing the annotation bias and visual homogeneity issues in existing CZSL datasets.

\item  We propose \methodname, a novel and efficient CLIP-based approach that removes the composition branch while and achieves state-of-the-art performance.
\end{itemize}


\section{\textcolor{black}{Related Work}}

\begin{figure*}[t!] 
    \centering
    \hspace{-4mm}
    \begin{subfigure}[b]{0.23\textwidth}
        \centering
        \includegraphics[width=\textwidth]{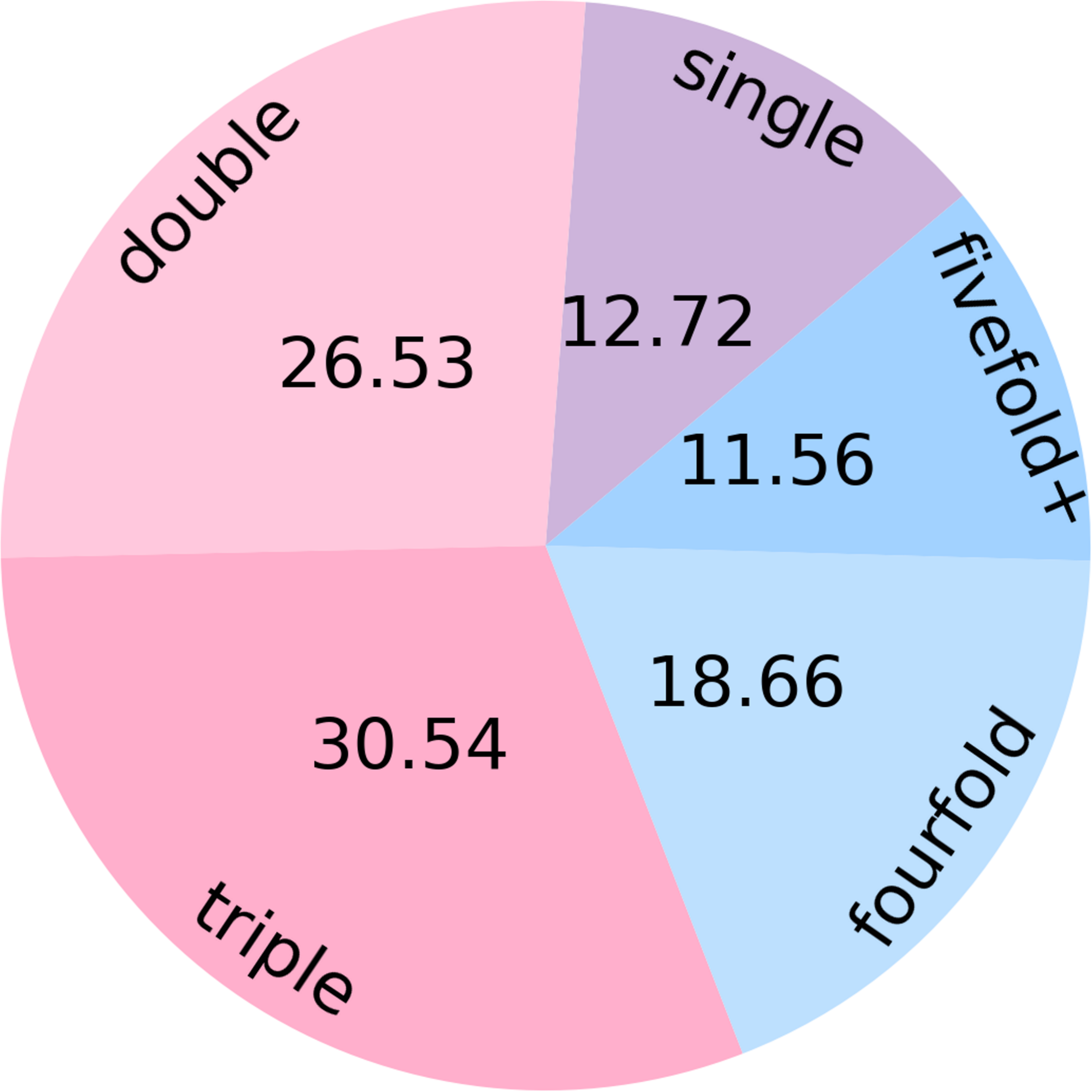}
        \caption{Label quantities.}
        \label{subfig:dataset_a}
    \end{subfigure}%
    \begin{subfigure}[b]{0.23\textwidth}
        \centering
        \includegraphics[width=\textwidth]{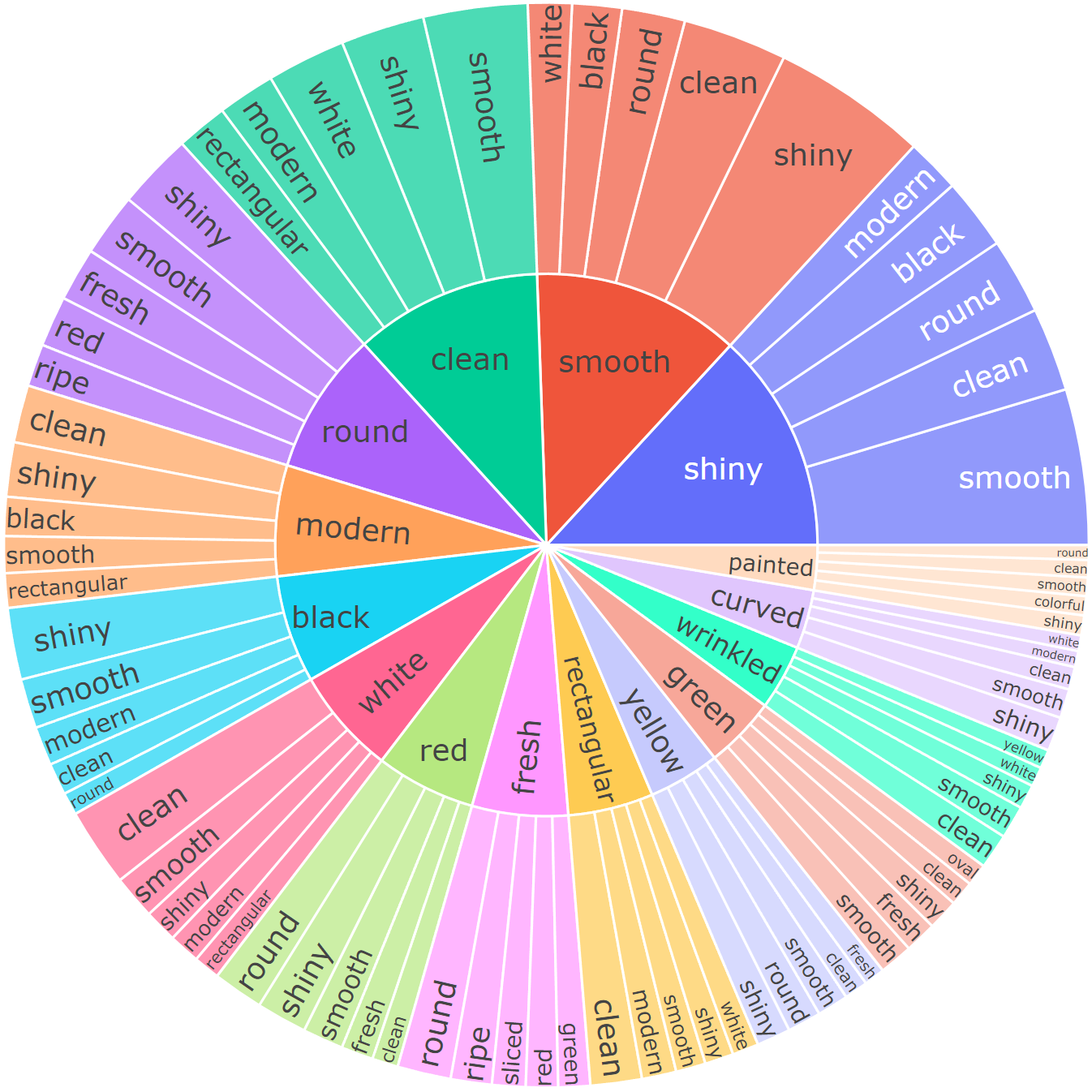}
        \caption{Attributes co-occurrence.}
        \label{subfig:b}
    \end{subfigure}
    \begin{subfigure}[b]{0.235\textwidth}
        \centering
        \includegraphics[width=\textwidth]{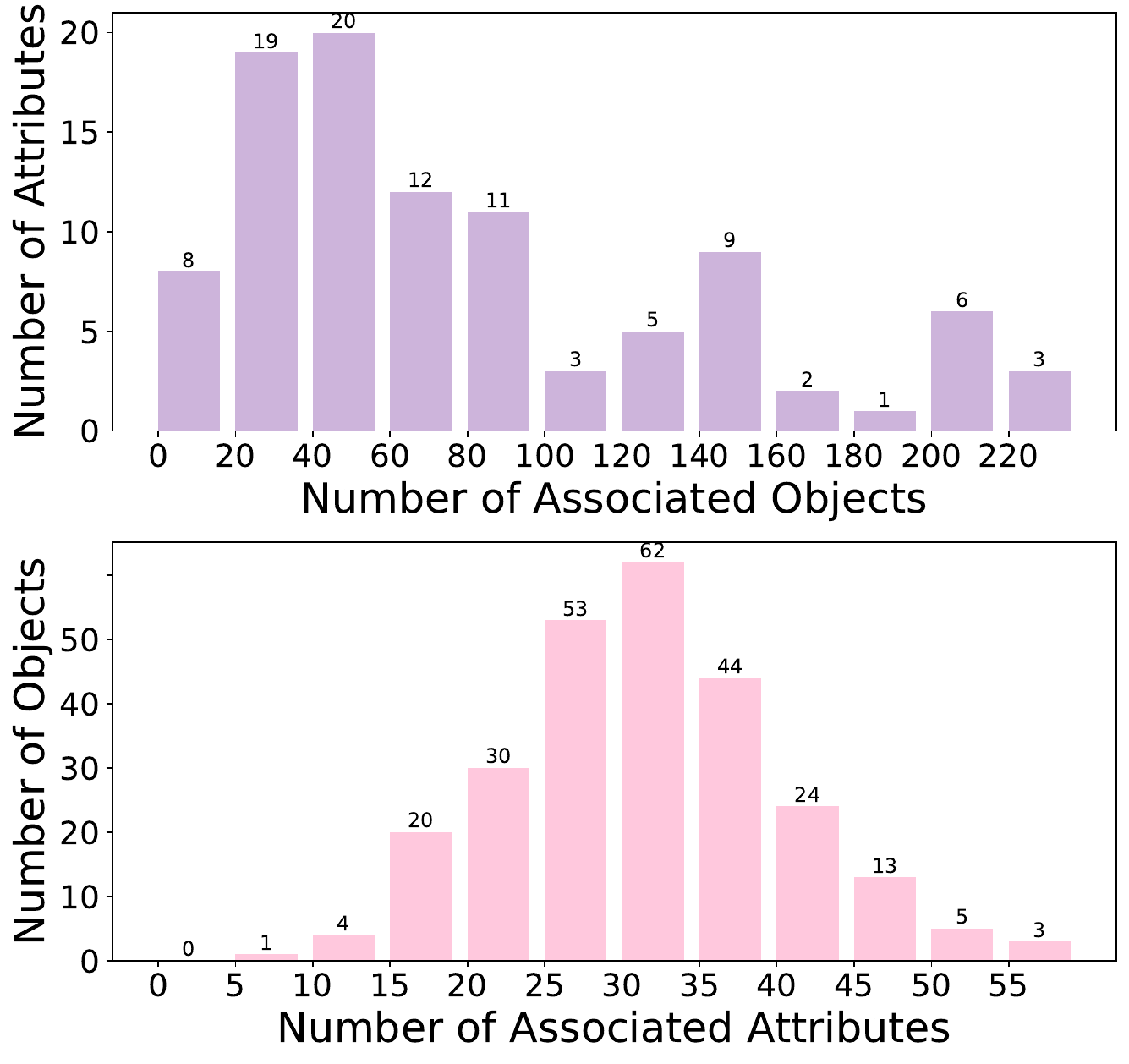}
        \caption{Binding relations.}
        \label{subfig:c}
    \end{subfigure}%
    \hspace{1mm}
    \begin{subfigure}[b]{0.285\textwidth}
        \centering
        \includegraphics[width=\textwidth]{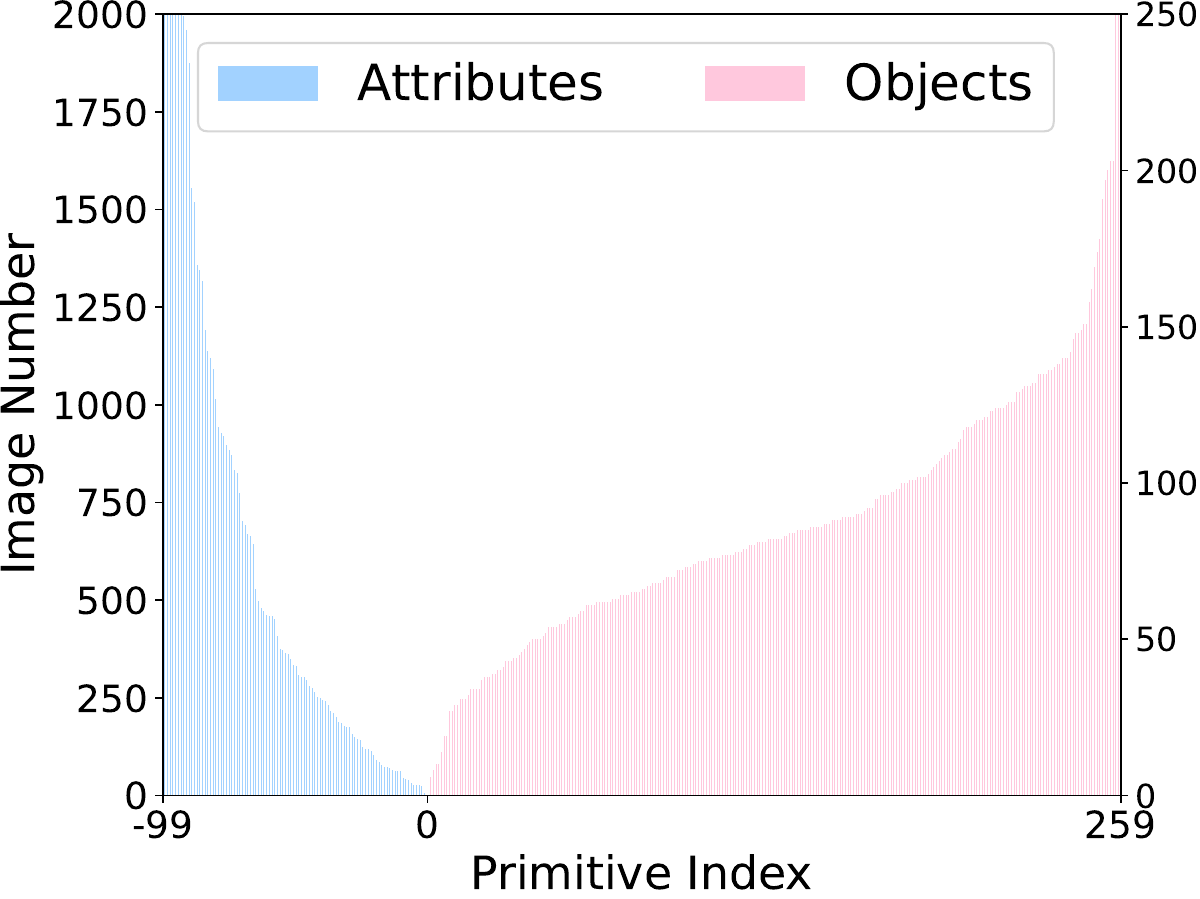}
        \caption{Instances per primitive.}
        \label{subfig:d}
    \end{subfigure}
    \caption{\textbf{Dataset statistics.} 
     (a) shows the proportion of images with different numbers of attribute labels;  (b) shows the co-occurrence of the top 15 attributes  with their most frequently associated attributes; (c) illustrates the binding relationships between attributes and objects. The top section displays the distribution of attributes across varying numbers of associated objects, while the bottom section  presents the reverse; (d) displays the number of images per primitive  for MAC. }
    \label{fig:dataset}
    \vspace{-4mm}
\end{figure*}

\begin{figure*}[ht!] 
    \centering
    \begin{subfigure}[b]{\textwidth}  
        \includegraphics[width=\textwidth]{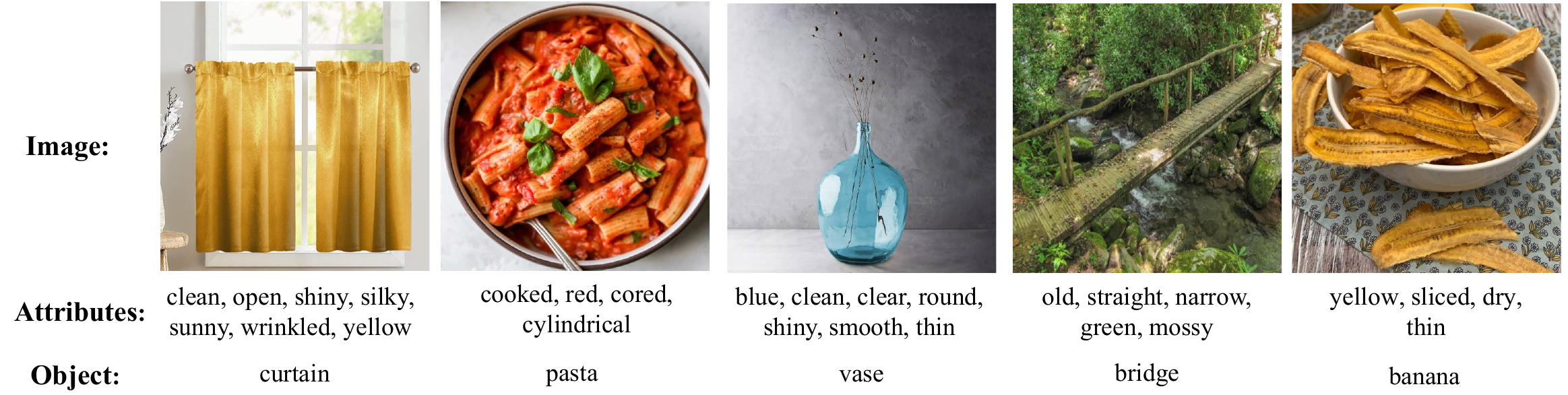}  
        \label{subfig:a}  
    \end{subfigure}  
    \vspace{-8mm}\caption{\textbf{Examples of MAC.} Our dataset provides comprehensive and representative attribute annotations for images.}
    \label{fig:samples}
    \vspace{-5mm}
\end{figure*}

\subsection{Compositional Zero-Shot Learning.}
Compositional Zero-Shot Learning (CZSL)~\cite{purushwalkam2019task, romera2015embarrassingly,  ruis2021independent, li2023compositional} aims to recognize unseen attribute-object compositions at test time while each semantic primitive exists in the training sample. 
\textcolor{black}{In contrast, conventional ZSL or GZSL methods~\cite{cheng2023discriminative,cheng2023hybrid,huang2024mrrfs}  typically focus on recognizing entirely unseen object classes. While these methods model relationships between local attribute features and object semantics, CZSL poses a more fine-grained scenario that requires understanding complex interactions between  attributes and objects.}
Existing methods for CZSL without large-scale pre-trained models can be divided into two main categories. One group of methods train two separate classifiers to predict attribute and object respectively~\cite{wang2023czslconditional,xu2022relation,Karthik2022KG-SP, ming2025adaptive, yang2024dual}. 
Another group of methods learns the joint attribute-object semantic representation of seen and unseen compositions using transformation functions in a shared feature space~\cite{anwaar2022leveraging,yang2023casual,jiang2024mutual, jing2024retrieval}. However, learning semantic primitives from scratch is difficult because of the complexity of the compositions. The recent work is starting to research how to transfer pre-trained vision-language models to composition prediction tasks~\cite{nayak2022learning, lu2022prompt, lu2023decomposed, huang2023troika,li2023compositional,li2023compositional,xu2024gipcol,bao2023prompting,shuang2025visual}. 
CSP~\cite{nayak2022learning} first adapts CLIP~\cite{radford2021learning} by replacing the text vocabulary with trainable tokens.
PromptCompVL~\cite{lu2022prompt} creates a completely learnable soft prompt including the prefix, state, and object. 
DFSP~\cite{lu2023decomposed} proposes a cross-modal decomposed fusion module to learn more expressive image features. 
Troika~\cite{huang2023troika} adopts a Multi-Path framework to dynamically adjust prompts and performs image-to-text feature fusion.
PLO~\cite{li2023compositional} solves composition predicting in a simple to complex manner by utilizing the knowledge of large language models. 
GIPCOL~\cite{xu2024gipcol} employs a graph convolutional network to enhance CLIP's word embeddings, creating text representations optimized for compositional learning.
In this work, we propose MVP-Integrator which eliminates the composition branch and utilizes self-attention for visual-primitive association. \textcolor{black}{This design echoes the core idea of mutual-assistance learning as discussed in \cite{xie2023mutual}, where task components are encouraged to interact and support each other for improved performance.}

\subsection{CZSL Datasets and Multi-attribute Datasets.} 
CZSL datasets like MIT-States~\cite{isola2015discovering}, UT-Zappos~\cite{yu2014fine}, and C-GQA~\cite{naeem2021learning} are commonly utilized for studying CZSL. 
UT-Zappos is a relatively simple dataset focusing on the composition of shoes, materials and brands.
MIT-States characterizes state variations within image classes using a diverse set of adjectives and their antonyms.
C-GQA includes numerous attributes and objects, derived from images cropped from the GQA~\cite{gqa} based on the bounding boxes.
The labels of these three datasets contain only one attribute. This misleads the model's understanding of attributes and results in inaccurate evaluation. 
Datasets such as COCO Attributes~\cite{patterson2016coco}, PACO~\cite{ramanathan2023paco}, and VAW~\cite{Pham2021VAW} offer multiple-attribute annotations. However, they do not guarantee comprehensive coverage of relevant attributes, and they exhibit trivial attribute-object relationships, with many objects being associated with a narrow set of attributes, limiting compositional complexity. 
\textcolor{black}{For example, our analysis of the large-scale VAW dataset, which contains 620 attribute classes and 2,260 object classes, reveals a sparse attribute-object mapping. We observe that over 40\% of its attributes are linked to fewer than 20 object classes, and 43.7\% of its object classes associated with five or fewer attribute classes. 
To tackle these problems, we construct a new dataset, MAC, a new dataset constructed specifically to compositional learning.  In contart to VAW, the proportions of sparse connections in MAC drop to just 8\% and 0\%. We achieved this diversity by collecting high-quality images with high visual variation for each object class.
We also ensure comprehensive attribute annotations through multiple strategies and propose a multiple attribute compositional learning task.}

\section{Multiple Attribute CZSL Benchmark}

\subsection{Dataset Overview}

The MAC dataset comprises 22,838 images and 17,627 compositions, with 99 attribute classes and 259 object classes. The MAC dataset is divided into training, validation, and testing sets with 14,591 , 3,007 and 5,240 images respectively. Our images encompass a variety of representative attributes and objects, with each object associated with multiple attributes. 

\Cref{subfig:dataset_a} shows the multi-attribute properties of MAC, with 60.76\% of the images annotated with three or more attributes. 
\Cref{subfig:b} illustrates the co-occurrence of different attributes. Frequently co-occurring attributes often share high semantic relevance, such as ``cooked'' and ``viscous'' or ``smooth'' and ``silky'' . Further analysis of the phenomenon is provided in section B of Appendix . 
\Cref{subfig:c} illustrates the complex relationship between attributes and objects in the MAC. 
The attributes in MAC are not exclusively linked to specific objects; 53\% of the attributes can applied to more than 60 classes of objects. The bottom part of \Cref{subfig:c} highlights a clear peak in the mid-range and reveals that 79\% of object classes  associated with more 25 attributes, demonstrating the rich appearance variations of images within the same object. 
\Cref{subfig:d} demonstrates number of instances per attribute and object. For object, 76\% of classes have image counts within one standard deviation of the mean, indicating a balanced dataset.

\subsection{Dataset Construction}

\begin{figure*}[!t]
    \centering    
    \begin{subfigure}[b]{0.57\textwidth}  
        \includegraphics[width=\textwidth]{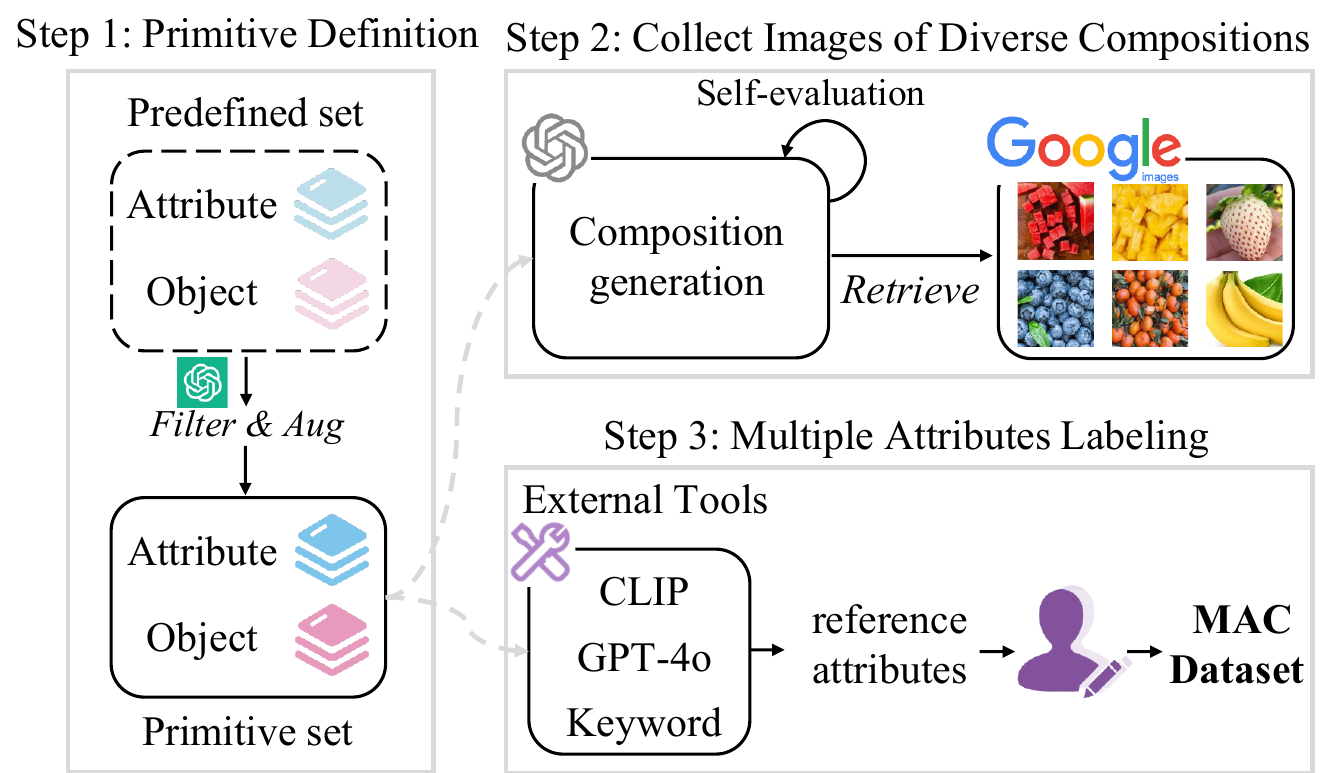}  
        \caption{Dataset construction}
        \label{subfig:constuct_a}  
    \end{subfigure}
        \begin{subfigure}[b]{0.35\textwidth}  
        \includegraphics[width=\textwidth]{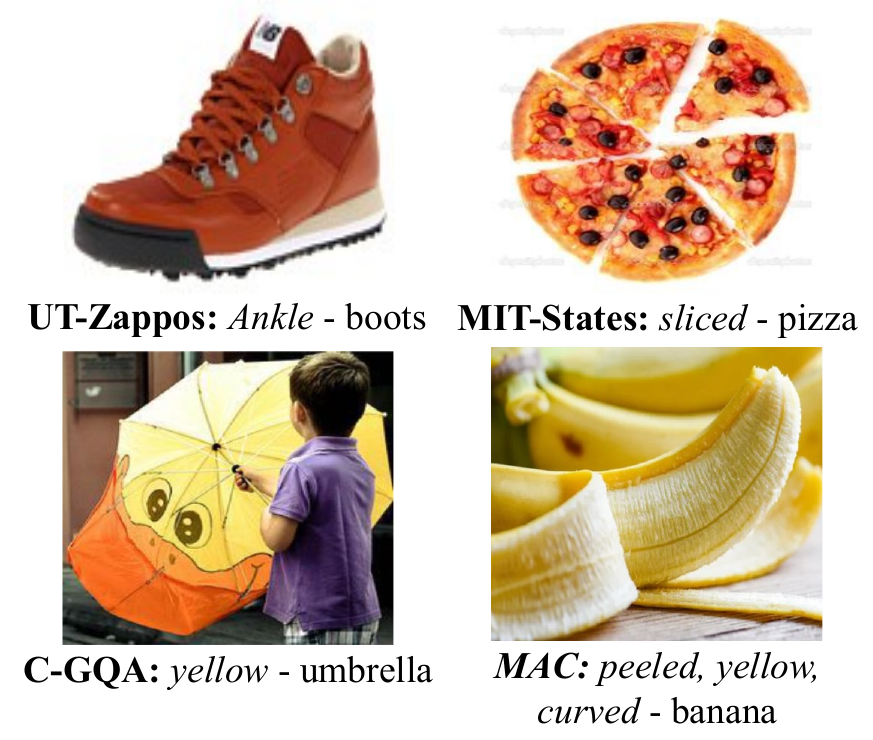}
        \caption{Samples from different datasets}
        \label{subfig:construct_b}  
    \end{subfigure}  
    \vspace{-2mm}
    \caption{(a) Step-by-step diagram of dataset construction. (b) Samples comparison of different datasets. The images from top left to bottom right are UT-Zappos~\cite{yu2014fine}, MIT-States~\cite{isola2015discovering}, C-GQA~\cite{naeem2021learning}, and MAC.}
    \label{fig:data_construction}
    \vspace{-5mm}
\end{figure*}

\begin{table}[t]  
\small
\tablestyle{5pt}{1.0}
\setlength\tabcolsep{7pt}
\caption{\textbf{Effectiveness of attribute candidates.} ``Attr Num'' means the total attribute number of each candidate set.
``Valid Proportion'' indicates the accuracy of attributes in each set, while ``Final Contribution'' represents their share in the final dataset.  The sum of ``Final Contribution'' exceeds 100\% due to overlapping attributes.}  
\begin{tabular}{l|cccc}   
\hline
Candidate Set & Attr Num & Valid Proportion  & Final Contribution \ \\ \hline\hline
Keyword        & 68,514  & 12.4\%     &  19.9\%     \\  
CLIP     &   114,190 &  14.8\%     &    31.1\%      \\  
GPT-4o     & 76,033  & 45.9\%       &   62.8\%         \\
Human Add   & 11,541 & 100.0\%     &      20.6\%        \\
\hline  
\end{tabular}  
\label{tab:label} 
\vspace{-5mm}
\end{table}  

\begin{figure}[t]
    \centering    
    \includegraphics[width=\columnwidth]{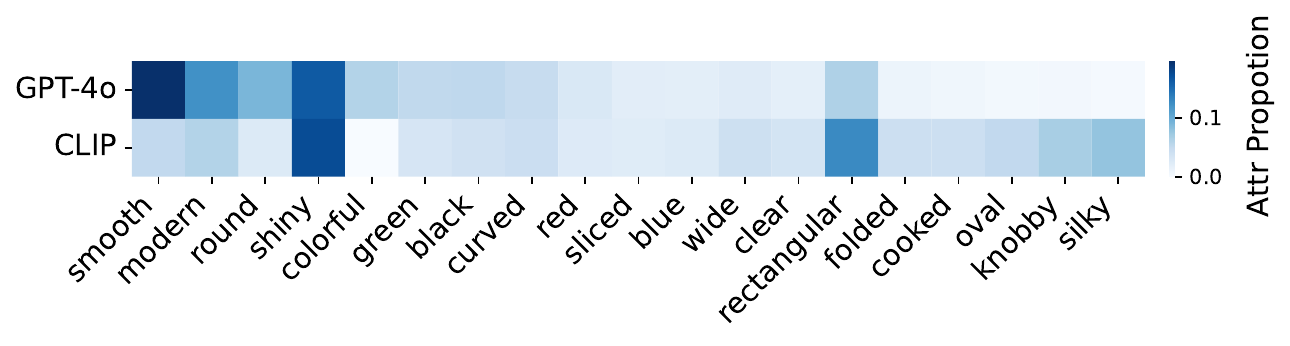}
    \vspace{-5mm}
    \caption{\textbf{Complementarity analysis.} The darker the color, the greater the proportion of all predictions.}
    \label{fig:tendency}
    \vspace{-5mm}
\end{figure}

In the MAC dataset, each image contains a single object label and multiple-attribute labels. To create a high-quality dataset with diverse images and comprehensive attribute annotations, we follow a three-step process as illustrated in \Cref{subfig:constuct_a}.
First, we carefully define the semantic primitives (\textit{i.e.}, attributes and objects) using GPT-4o and manual verification (\textbf{Primitive Set Definition}). We use GPT-4o to generate diverse compositions for each object by selecting suitable attributes. These compositions serve as keywords to retrieve images from search engines, forming the initial dataset (\textbf{Collect Images of Diverse Compositions}).
To cover suitable attributes as comprehensive as possible, we use CLIP~\cite{radford2021learning} and GPT-4o to predict attributes for each image. Annotators then select correct attributes for each image based on these predictions and the initial search keywords (\textbf{Multiple Attributes Labeling}).
This process results in a high-quality multi-attribute composition dataset with comprehensive attribute labels. 

\textbf{Primitive Set Definition.}
The primitive set defines the object classes and attribute classes, which directly determines the dataset's scope and significantly influences its quality. Existing datasets, such as MIT-States~\cite{isola2015discovering} and C-GQA~\cite{naeem2021learning}, often include numerous synonyms and non-visual attributes, hindering the learning of semantic primitives and compromising evaluation accuracy.  We carefully define the primitive set by leveraging GPT-4o. 
Take the attribute set as an example, we consider the attribute set of MIT-States as an initial set and refine it through  following steps. 
First, we ask GPT-4o to eliminate attributes that are not visible. Then, we categorize the remaining attributes into mutually exclusive groups and merge synonyms. Finally, we ask GPT-4o to enrich each group with new attributes.
To ensure that each attribute has a unique meaning, we merge synonyms using WordNet~\cite{miller1990introduction} and  manually check  the final primitive set. 
The object set is constructed in the same manner. The resulting primitive set contains 99 attributes and 259 objects. 

\textbf{Collect Images of Diverse Compositions.}
For a high-quality composition dataset, the diversity of attributes associated with each object is crucial. It is not enough to simply collect a large number of images for each object type. These images must show the object with different attributes. This diversity is beneficial for learning how attributes and objects interact and preventing the models from inferring attributes solely based on object classes.
To achieve this, we create a diverse set of feasible attribute-object compositions. 
Specifically, we use GPT-4o to generate plausible compositions that include three attributes each.
Compared to single attribute compositions, this strategy generates more diverse compositions with attributes from different descriptive perspectives. We also leverage GPT-4o to perform self-evaluation, rating each composition's logical validity and feasibility on a scale from 1 to 5. Compositions with scores above 3 are selected as the final composition set.
We use these compositions as keywords to crawl images from Google. Leveraging the search engine's robust retrieval capabilities, we are able to collect diverse images.

\textbf{Multiple Attributes Labeling.}
To address the challenge of comprehensive multi-attribute annotation, our labeling process provides annotators with three reference sets: a keyword set (from image collection), a GPT-4o set, and a CLIP set.
The keyword set consists of search keywords used for image collection, containing three attributes and one object per image. These terms reflect data from search engines (like tags and file names), providing a relevant starting point for annotation. However, they may not always accurately represent the image content, as illustrated by the term "frozen" potentially retrieving images from the movie instead of literally frozen objects.
The GPT-4o set is generated by GPT-4o, providing more accurate attribute candidates. However, it has hallucinations, occasionally assigning textually plausible but visually incorrect attributes (e.g., predicting a frozen apple as ``fresh'').
The CLIP set is obtained by using CLIP to retrieve the top-5 attributes for a given image and its object category.
As shown in \Cref{fig:tendency}, the CLIP set and the GPT-4o set exhibit a distinct complementary relationship. Using both sets ensures more comprehensive attribute annotation.
Furthermore, annotators are instructed to supplement other appropriate attributes beyond the three sets and  discard images  without a prominent object or of poor quality. \Cref{tab:label} shows the proportion of valid attributes in each set and their contribution to the final annotations.

\subsection{Annotation Quality and Potential Data Leakage}
\begin{figure}[t]
    \centering
    \includegraphics[width=1.0\columnwidth]{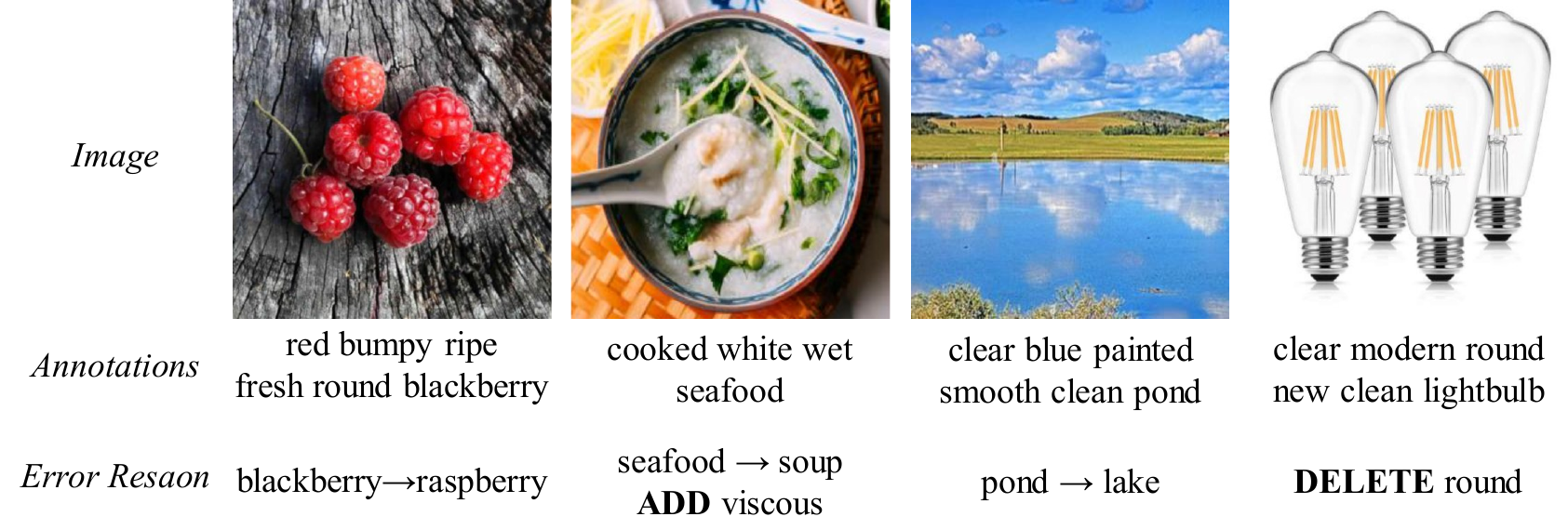}
    \vspace{-5mm}
    \caption{\textbf{Examples of annotation errors.} Selected from MVP-Integrator's top-1 prediction failures, which tend to reveal underlying annotation errors.  ``$\rightarrow$'' indicates an object correction. ``ADD'' denotes a missing attribute. ``DELETE'' denotes an incorrect attribute. } 
    \label{fig:error_annotations}
    \vspace{-5mm}
\end{figure}

\textcolor{black}{
\textbf{Annotation Quality.}
To assess the quality of our annotations, we conducted an error analysis on carefully selected samples. Representative examples of error cases are shown in \Cref{fig:error_annotations}.
We used our trained MVP-Integrator model to find 200 instances where the top-1 predictions didn't match the annotations. These instances tended to deviate from the main distribution and are more likely to reflect annotation issues. 
Our inspection revealed only 7 annotation errors, which we classified into three types: Object Confusion (4 cases), referring to cases where the object was mislabeled as a visually similar or ambiguous one; Missing Attributes (2 cases), where valid attributes were present in the image but not annotated; and Incorrect Attributes (1 case), where the annotated attributes did not align with the actual visual content. The majority of these errors were due to object confusion. Considering that these 200 samples represent the most error-prone subset of the entire dataset, and given the inherent difficulty of comprehensive multi-attribute annotation, this low error rate demonstrates the high quality of our dataset.
}
  
\textcolor{black}{\textbf{Data Leakage Analysis.} We've also conducted a thorough analysis to address concerns about data leakage from using CLIP in both annotation and training. First, we used diverse strategies for annotations stage to prevent reliance on any single source. As shown in ~\Cref{tab:label},  CLIP suggestions made up only 30.9\% of the final labels, so our dataset isn't dominated by its biases.
Second, we empirically verify that our model does not disproportionately benefit from CLIP-sourced labels. Only 16.8\% of the top-1 correct predictions corresponded to attributes uniquely suggested by CLIP, compared to 30.3\% from GPT-4o, showing the model isn't just memorizing CLIP-derived labels.
Finally, we performed an ablation study by retraining our model with all raw CLIP attribute suggestions. The results showed a significant performance decline, with the Exact Match score dropping from 17.42\% to 10.81\%. This degradation confirms that uncurated CLIP data introduces noise and doesn't improve learning, further validating that our dataset construction process doesn't lead to data leakage.}

\subsection{Task Formulation}
\label{sec:task}
Given the attribute set $A = \{a_1,a_2,\cdots,a_{|A|}\}$ and object set $O = \{ o_1,o_2,\cdots,o_{|O|} \}$ as the semantic primitives, where $|\cdot|$ denotes the number of elements in the set, the multi-attribute compositional label space is defined as
   $C = \left\{ \left\langle S, o \right\rangle \mid S \subseteq A, o \in O \right\}$. We define the set of seen compositions and unseen compositions as $C^s \subseteq C$ and $C^u \subseteq C$, respectively,  where $C^s \cap C^u = \emptyset$. During training, only the data of seen compositions is provided.
Implementing Image-to-Text retrieval for multi-attribute compositions is impractical due to the massive search space. Therefore, we design a multi-label single attribute composition classification task for our dataset.  This task aims to predict compositions in the test space $C^t$ in both closed-world setting and open-world setting. 
In closed-world setting, the solution space is defined as $C^t = \left\{ \langle a, o \rangle \mid \exists \langle S, o \rangle \in  C^s \cup C^u, \, a \in S \right\}$. Additionally, we focus more on the open-world setting because it is more consistent with real-world scenarios. In this setting, the solution space is defined as the Cartesian product of the attribute set and the object set, \textit{i.e.},  $C^t = A \times O$.

\subsection{Evaluation Metric}
\label{sec:metric}
To evaluate the accuracy and comprehensiveness of model respectively, we use \textbf{Exact Match}, Top-1 precision (\textbf{Top1-P}), Top-5 Recall (\textbf{Top5-R}) and \textbf{Coverage} following ~\cite{wu2017unified,garcia2024thorough}. 
Exact Match is the proportion of instances that all true labels are ranked higher than any false labels.
The Top1-P reflects the model's ability to correctly predict compositions. 
The Top5-R evaluates the ability of a model to identify relevant labels within its top 5 predictions.  
Coverage quantifies the model's ability to capture all true labels by computing the smallest integer $K$ such that the top $K$ predicitons cover all true labels of an instance.
We also introduce Top-1 Precision for attributes (\textbf{Top1-P-attr}) and objects (\textbf{Top1-P-obj}) to measure the model's performance on individual primitives. 
To ensure consistency with prior work, we also report results on \textbf{AUC}, \textbf{Seen}, and \textbf{Unseen} following ~\cite{naeem2021learning}. We use the top1-P to compute seen and unseen accuracies.


\section{Proposed Method}

\label{sec:method}

\begin{figure*}[t!]
    \centering    
    \includegraphics[width=0.95\textwidth]{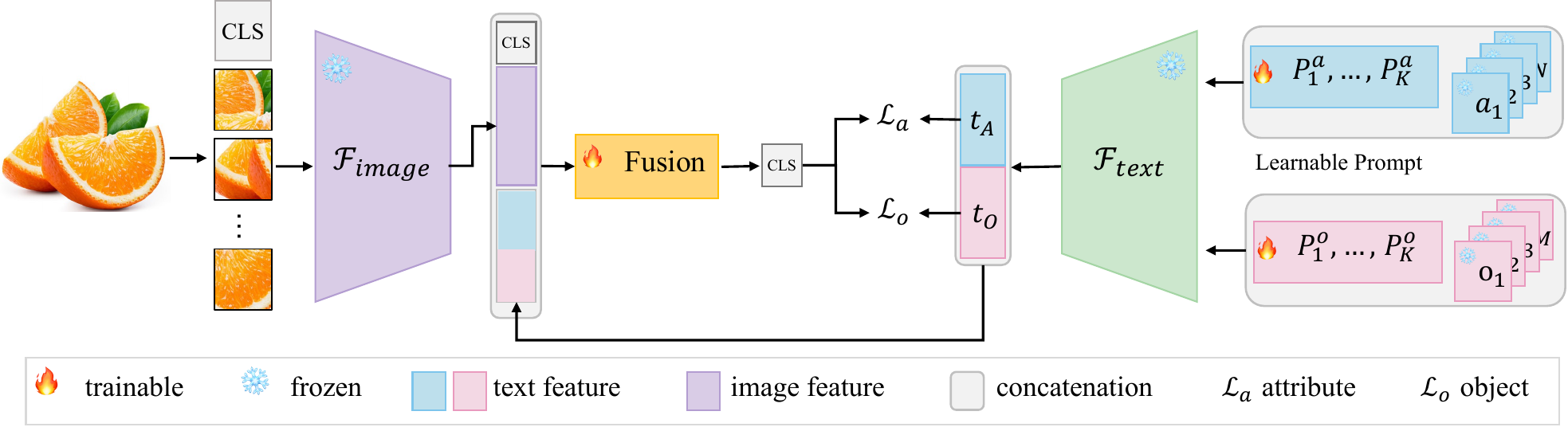}
    \caption{\textbf{The proposed method.} 
    $\mathcal{L}_a$ is BCE loss on attribute classification and $\mathcal{L}_o$ is Cross-Entropy loss for object classification. }
    \label{fig:method}
    \vspace{-5mm}
\end{figure*}

Compared to previous benchmarks, multi-attribute compositional learning is more challenging as it requires the model to identify all feasible attributes. 
To achieve accurate composition predictions, a model must disentangle semantic primitives to independently learn attributes and objects from composition data. It should also effectively capture relationships between primitives, including attribute-attribute and attribute-object associations.
However, existing models struggle with multi-attribute composition learning since they are designed for single-attribute datasets, ignoring the relationship between different attributes.
We propose \textbf{M}ulti-attreibute \textbf{V}isual-\textbf{P}rimitive \textbf{I}ntegrator (\methodname) to adapt CLIP~\cite{radford2021learning} for multi-attribute composition learning, as shown in \Cref{fig:method}. 
\methodname is \textit{the first CLIP-based CZSL model to eliminate compositional inputs} and instead uses dual-branch prompt tuning for primitive disentanglement and leverages self-attention to capture relationships between primitives automatically.

\textbf{Prompt Tuning.}  
Unlike previous methods, we discard compositional text inputs and instead employ a dual-branch prompt tuning strategy with independent learnable prompts for attributes and objects. For an attribute-object composition $(a\;o)$, we convert the natural language prompt ``a photo of ${a}$ ${o}$'' into two independent prompt:  $P_a = [p^a_1 \; p^a_2 \; \dots \; p^a_K \; v_a]$ and $P_o = [p^o_1  \; p^o_2 \; \dots \; p^o_K \; v_o]$ , where $[p_1 \;  p_2  \; \dots \;p_K]$ are the trainable embedding and $v_a , v_o$ are the frozen word vocabulary embeddings of $a,o$. The trainable prompts are fed into the text encoder $\mathcal{F}_{text}$ to obtain the text representation of all primitives, formulated as
\begin{equation}
\begin{aligned}
    t_{a_i} &= \mathcal{F}_{\text{text}}(P_{a_i}), \quad
    t_{o_j} = \mathcal{F}_{\text{text}}(P_{o_j}), \\
    t_A &= [t_{a_1}\; \ldots\; t_{a_{|A|}}], \quad
    t_O = [t_{o_1}\; \ldots\; t_{o_{|O|}}].
\end{aligned}
\end{equation}

\textbf{Multi-Attribute Visual-Primitive Association.}  
Due to the complexity of the compositional space (the cartesian product of the attribute and object sets), existing methods often require specialized modules for attribute-object association and separate mechanisms for text-image feature fusion.
In contrast, our method eliminate the compositional branch significantly reducing the complexity of text features. This enables us to employ a lightweight transformer encoder to jointly model relationships between attributes and objects, inter-attribute correlations, and cross-modal interactions between vision and language.
Specifically, let $I$ denote the concatenated image patch representations and $C$ denote the class token representation extracted by the CLIP's visual encoder. $T = [t_A;t_O]$ represents the concatenated text representations extracted by CLIP's text encoder for all attributes and objects. The combined tokens $M = [C;I;T]$ are fed into a lightweight transformer encoder to model both the relationship between different semantic primitives (attributes and objects) and the relationship between the image and the text. Finally, the cosine similarity is calculated between the refined image class token $C'$ and the original text token representations $T$.
It's worth noting that the methods without a composition branch do not require training on all possible compositions. It significantly reduces the computational resources for training and inference on the combined dataset: the input number to the text encoder drops from $|A|\times|O|$ to $|A| + |O|$.

\textbf{Training objective.}
As shown in \Cref{fig:method}, MVP-Integrator contains two  branches for attributes and objects respectively. Given the text representation for all primitives and visual features, the logits of of assigning labels of the attribute $a_i$, object $o_j$ can be computed separately as
\begin{equation}
s_{x,a_i} = \frac{\mathbf{C}'_x \cdot \mathbf{t}_{a_i}}{||\mathbf{C}'_x|| \cdot ||\mathbf{t}_{a_i}||}, \quad s_{x,o_j} = \frac{\mathbf{C}'_x \cdot \mathbf{t}_{o_j}}{||\mathbf{C}'x|| \cdot ||\mathbf{t}{o_j}||}.
\end{equation}
The attribute branch is trained with BCE loss, formulated as:
\begin{equation}
\begin{aligned}
\mathcal{L}_a = - \frac{1}{|\mathcal{X}|} \sum_{x\in \mathcal{X}} \sum_{i=1}^{|A|} \big[
& y_{xi}\log \sigma(s_{x,a_i}) \\
& + (1 - y_{xi}) \log(1 - \sigma(s_{x,a_i})) \big],
\end{aligned}
\end{equation}
where $y_{xi}\in{0,1}$  indicates the presence of attribute $a_i$ of image $x$, and $\sigma(\cdot)$ is the sigmoid activation function. The object branch is trained using the standard Cross-Entropy (CE) loss:
\begin{equation}
    \mathcal{L}_o = - \frac{1}{|\mathcal{X}|} \sum_{x \in \mathcal{X}} \log \left( \frac{\exp(s_{x,o_{y_x}})}{\sum_{j=1}^{|O|} \exp(s_{x,o_j})} \right),
\end{equation}
where $o_{y_x}$ denotes the labeled  object class for image $x$. Therefore, the overall training loss $\mathcal{L}$ is defined as:
\begin{equation}
    \mathcal{L} = \mathcal{L}_a + \mathcal{L}_o.
\end{equation}

\section{Experiments}

\input{tabs/table_main_result}

\subsection{Implementation Details}
We evaluate methods on the multi-label single composition task described in \Cref{sec:task} with metrics in \Cref{sec:metric}. CGE uses a ResNet18 pretrained on ImageNet, while other methods use a CLIP ViT-L/14 backbone. Troika is trained and evaluated on an NVIDIA L20 GPU, whereas other methods use an RTX 4090 GPU. Hyperparameter details are provided in  section D of Appendix. All the methods are trained and evaluated in the same way. During training, we use cross-entropy loss for the object branch and BCE loss for the multi-label branch (attribute branch and composition branch). During inference, we add all the predictions to get the final composition prediction:
\begin{equation}
 \tilde{P}((a,o)|x_i) =  P(a|x_i) + P(o|x_i) + P((a,o)|x_i).
\end{equation}
These probability distributions are derived by applying the softmax function to the respective image-text similarity scores.
For methods that lack one or more branches, we disregard the missing branches during both training and inference. We do not use any special post processing (e.g., feasibility masking \cite{mancini2021open}) for open-world setting to directly measure the model's compositional learning ability.
We also evaluate visual-language models (VLMs) like LLaVA~\cite{liu2024llava}. We use a probabilistic inference technique to fully utilize their potential. 
We calculate class probabilities by summing the token probabilities for each class name and rank \textit{prob(class name|image, prompt)} as the final prediction~\cite{brown2020GPT3,zhang2024whyVLM}.
Calculating these probabilities directly in the composition space is infeasible. Instead, we compute the probabilities for all the classes of objects and attributes for each image and sum them to derive the composition probability.

\subsection{Main Results}

We experiment seven models on MAC. CGE~\cite{naeem2021learning} is the only one not based on CLIP. CLIP~\cite{radford2021learning}, CoOP~\cite{zhou2022learning}, CSP~\cite{nayak2022learning}, DFSP~\cite{lu2023decomposed}, GIPCOL~\cite{xu2024gipcol}, and Troika~\cite{huang2023troika} are all common CZSL baselines based on the CLIP ViT-L/14 backbone. LLaVA~\cite{liu2024llava} is the most widely used VLM, introducing more parameters compared to other models. 

\Cref{tab:main_result} shows the performance of various models in both the closed-world and open-world settings on the MAC dataset. In the \textbf{closed-world} setting, \methodname significantly outperforms other methods, achieving 17.42\% Exact Match, 63.45\% Top1-P, 62.37\% Top5-R and 27.89 on AUC. In the \textbf{open-world} setting, our method also achieves the best results. Notably, \methodname maintains consistent performance across both settings, while other methods show a significant drop in the open-world setting.
LLaVA has high prediction accuracy but performs poorly in overall prediction quality, as measured by the Top5-R and Coverage.
Compared with the closed-world setting, the open-world setting lacks infeasible composition priors, making it more challenging and better aligned with real-world applications.
Methods based on composition branch show a significant drop in performance from closed-world to open-world settings, with a relative 22.01\% drop in Exact Match. 
Specifically, the performance of models based on CLIP in terms of the Top1-P-attr drops by at least 3.98\%. This indicates that existing methods are heavily influenced by composition priors and have a weaker understanding of the essential characteristics of attributes. 
In contrast, our \methodname shows a difference of only relatively 3.7\% in Exact Match and 0.63\% in  Top1-P-attr between different settings, with overall results remaining nearly unchanged. 

\subsection{Efficiency analysis}

\input{tabs/tab_efficiency}

\begin{figure*}[t]
    \centering    
    \includegraphics[width=1.0\textwidth]{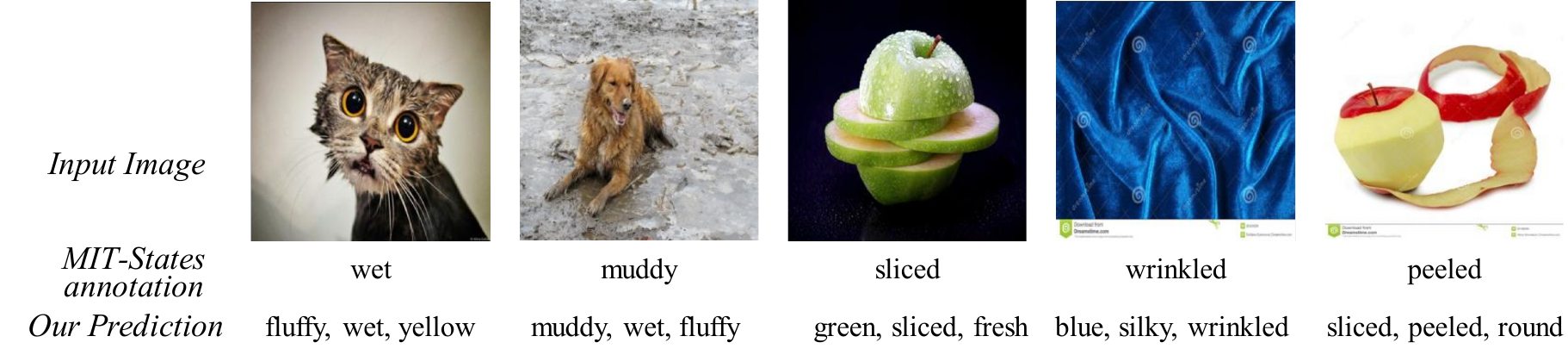}
    \vspace{-5mm}
    \caption{\textbf{Qualitative results for cross-dataset evaluation. }
    We directly test the MAC-trained MVP-Integrator on MIT-States~\cite{isola2015discovering} data, finding that its top-3 attribute predictions introduce suitable new attributes while retaining the original ground truth.
    }

    \label{fig:qual}
    \vspace{-5mm}
\end{figure*}

\textcolor{black}{
As shown in \Cref{tab:efficiency}, the proposed \methodname demonstrates superior efficiency in the open-world setting. Compared to existing methods, it is  approximately 25$\times$ faster for inference and requires only about 15\% of the GPU memory.
This  significant improvement comes  from a fundamental architectural innovation that addresses a computational bottleneck present in prior work. Previous CLIP-based methods rely on a composition branch to get the logits of every possible attribute-object pair. This approach leads to a computational complexity that scales quadratically with the number of primitives ($|A|\times|O|$).
Their primary bottleneck is the need to generate a unique text embedding for every single one of these potential compositions. This requires passing all $|A| \times |O|$ compositions through the  CLIP text encoder before performing cross-modal fusion, resulting in prohibitively high latency and substantial memory consumption.}
\textcolor{black}{
In contrast, \methodname completely eliminates the composition branch.  The core of our efficiency comes from drastically reducing the workload of the CLIP text encoder. MVP-Integrator only needs to compute text embeddings for all attributes and objects, resulting in a total of  $|A| + |O|$ text encoder computations.  This  design is the direct cause of its superior efficiency, making \methodname a far more scalable and practical solution for real-world applications with much more primitives.}

\subsection{Cross-dataset Evaluation}

\input{tabs/tab_cross}

To prove that MAC is better for the compositional learning, we conduct cross-dataset testing on four models. We train these models on MAC and evaluate them on a subset of MIT-States. 
All models are trained on MAC and evaluated on MIT-States through our aligned subset (4,982 images covering 70 attributes and 134 objects shared between datasets).
Given the vocabulary differences between MAC and MIT-States, we create an aligned subset of 4,982 MIT-States test images covering 70 attributes and 134 objects with MAC.
We train Troika, \methodname, CoOP, and Tri-CoOP on MAC for 40 epochs and evaluated them on this intersection set. Specifically, Tri-CoOP extends CoOP by adding classification branch for attributes and objects. We directly compare Top-1 pair accuracy for intuitively comparison.

\Cref{tab:cross} shows the results of cross-dataset evaluation. We can observe that the models trained in MAC achieve comparable results with their counterparts trained on MIT-States, except for Troika. This discrepancy likely stems from Troika's use of Adapter~\cite{houlsby2019parameter} to finetune CLIP's visual encoder, which may have reduced its generalization ability.
The cross-dataset performance of \methodname and Tri-CoOP even outperforms their counterparts. This indicates that the multiple comprehensive attribute labels of MAC help to improve the quality of the model’s composition predictions. 
In \Cref{fig:qual}, we use \methodname trained on the MAC to generate the top three predictions on images from the MIT-States. \methodname provides representative and relevant new attributes while preserving the original annotations in its top predictions.
These newly identified attributes provide a more detailed description  compared to the single-attribute annotation in MIT-States.
Furthermore, although the model's predictions are reasonable, they are often misjudged as incorrect due to the single-attribute annotation of MIT-States. This underscores how single attribute annotations can lead to inaccurate evaluation.

\subsection{Ablation Studies}

In \Cref{tab:ablation}, we conduct extensive experiments and report the closed-world results to show the effectiveness of our design.

\input{tabs/tab_ablation}

\noindent\textbf{Disentangling Primitives.} \textcolor{black}{ Group 3, which uses separate prompt tuning branches for attributes and objects, significantly outperforms Group 2, which uses a single composition branch. This improvement underscores the benefit of disentangling primitives, showing that predicting attributes and objects separately is more effective than predicting compositions. 
To further support this finding, we conducted an additional experiment. When we added a composition branch (trained similarly to CoOp [1]) to our full MVP-Integrator, the performance dropped considerably, with the Exact Match score falling from 17.42\% to 9.26\% and the AUC from 27.89\% to 24.65\%. This reinforces our conclusion that a dual-branch structure is superior for this task and suggests that effectively leveraging a composition branch for multi-attribute compositional zero-shot learning remains an valuable research question.}

\noindent\textbf{Multi-attribute Visual-Primitive Association.} The integration of the Cross-Modal Fusion module  results in an obvious improvement in Exact Match compared to the models without it \textcolor{black}{(Group 4 vs. Group 3)}. 
This highlights the importance of  modeling the  interactions between images and language and relationships across different primitives. \textcolor{black}{We further analyze the effect of primitive relationships in ~\Cref{sec:attention_analysis}.}

\noindent\textbf{CLS Token vs. Patch Token.} 
\textcolor{black}{To understand the source of visual information for our Integrator module, we compare two variants. Group 4 uses the single CLS token from CLIP's image encoder. Group 5 uses the patch token and average the Integrator's output tokens for the final prediction. 
Group 5 performs worse than Group 4, especially in terms of attribute  accuracy. 
This indicates that the CLS token  plays a more significant role in enhancing the image features, as it has been directly trained to align with text features.}

\noindent\textbf{Patch Token Contributions.} \textcolor{black}{While the [CLS] token is crucial, local features from patch tokens also provide valuable information. Group 7 (full model) outperforms Group 4 in all three metrics. This indicates that Patch Token features  enhance the model's ability to understand semantic primitives.  As noted in MaskCLIP~\cite{zhou2022maskCLIP}, patch tokens retain local semantic details, which can help the model better understand fine-grained local attributes such as ``dented'' or ``spiky''.}




\subsection{Attention Analysis}
\label{sec:attention_analysis}
\begin{figure}[t!]
    \centering
    \includegraphics[width=0.9\columnwidth]{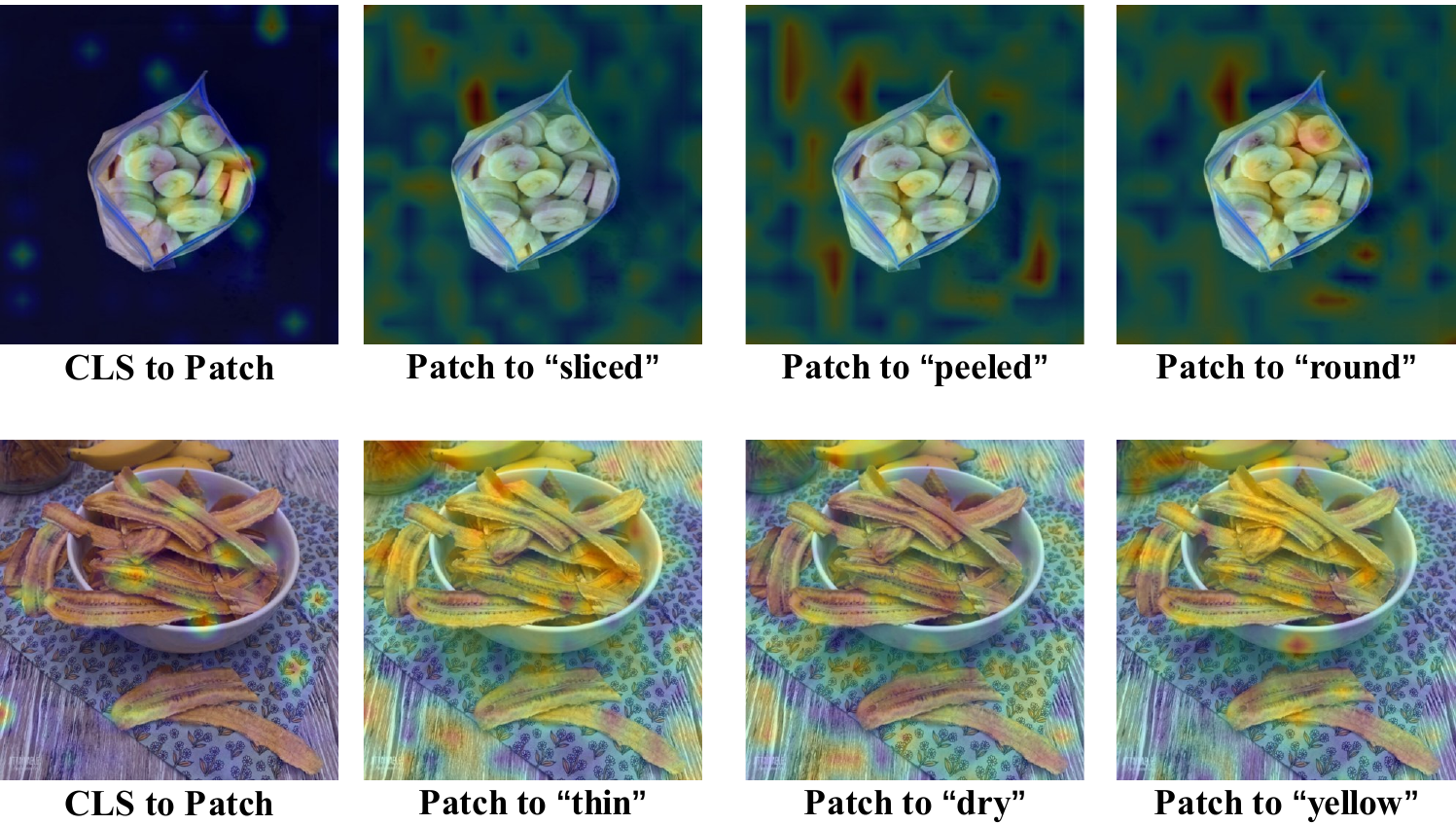}
    \caption{\textbf{Attention visualization.} ``CLS to Patch'' visualizes the attention from the image's CLS token to the patch tokens, highlighting salient visual regions. The subsequent columns (``Patch to Attribute'') visualize the attention from the image local region tokens  to corresponding attribute text tokens.  } 
    \label{fig:attention_viz}
\vspace{-3mm}
\end{figure}
\textcolor{black}{
In \Cref{fig:attention_viz}, we conducted an attention visualization analysis to better understand the specific mechanism of the image patch tokens within our Integrator module. 
The attention maps of ``Patch to Attribute'' tend to exhibit a fixed pattern across different samples.  They do not show a clear correspondence between each specific attribute and its associated image region. 
This is because patch tokens are not directly supervised for attribute alignment. This behavior is similar to that observed in GATv2\cite{brody2022GATv2} on fully connected graphs.}

\textcolor{black}{
Despite the lack of explicit patch-level attribute alignment, the attention maps still reveal meaningful  behavior of patch token. 
The ``CLS to Patch'' maps show the CLS token focusing on salient visual regions, such as the sliced bananas in the bag or the texture of the dried banana chips. This indicates that the main role of patch tokens is to provide rich local details to the global CLS token, which then performs the final prediction. While the model may not learn intuitive patch-level attribute alignment, it still effectively extracts and utilizes useful visual information from local regions.}
\begin{table}[t!]
\tablestyle{5pt}{1.0}
\setlength\tabcolsep{10pt}
\centering
\caption{  \textbf{Impact of primitive interactions within the Integrator module.} Results are obtained by retraining the model with specific attention interactions disabled: “Attr–Obj” denotes attention between attributes and objects, “Attr–Attr” denotes attention among attributes, and “All-Primitives” denotes attention over all primitive tokens.   }
\label{tab:attention_mask_ablation}
\begin{tabular}{l|ccc}
\hline
\textbf{Model} & \textbf{Exact Match} & \textbf{Attr} & \textbf{Obj} \\ \hline
Ours & 17.42 & 71.55 & 86.45 \\ \hline
Ours w/o Attr-Obj & 15.25 & 69.04 & 86.87 \\
Ours w/o Attr-Attr & 16.97 & 70.67 & 86.68  \\ 
Ours w/o All-Primitives & 15.11 & 69.15 & 85.78 \\ \hline
\end{tabular}
\vspace{-5mm}
\end{table}

\textcolor{black}{
We also conducted an attention masking experiment to measure the impact of primitive relationships. As shown in \Cref{tab:attention_mask_ablation}, masking the interaction between attribute and object tokens (Ours w/o Attr-Obj) caused a significant drop in Exact Match and attribute accuracy. This drop was more pronounced than when we masked intra-attribute attention (Ours w/o Attr-Attr). This confirms that modeling the relationship between attributes and objects is more critical to our model's performance than modeling the relationships among attributes themselves.}

\section{Conclusion}

In this paper, we address the limitations of the existing compositional zero-shot learning (CZSL) dataset and introduce the Multi-Attribute Composition (MAC) dataset, which focuses on multi-attribute composition prediction. Unlike existing datasets with single attribute annotations, MAC offers comprehensive and representative attribute annotations. It contains a large number of compositions with multiple attributes, allowing for robust evaluation of compositional learning. 
We evaluate nine  methods on MAC and propose \methodname as a strong baseline.
Our work contributes to the  compositional learning by highlighting the significance of multi-attribute compositions in real-world scenarios. MAC dataset provides a valuable resource for further research  in compositional learning tasks. Future work could explore the fine-grained correspondence between images and multiple attributes.
{\small
\bibliographystyle{IEEEtran}
\bibliography{IEEEabrv,main}
}

\end{document}

%% file: tabs/tab_dataset.tex
\begin{table*}[t]
\centering
\small
\caption{\textbf{Comparison of existing CSZL datasets.} ``Avg.attr'' represents the average number of attribute types per object type, while ``Avg.obj'' denotes the average number of object types per attribute type.}
\setlength\tabcolsep{5.5pt}
\begin{tabular}{c|ccccccccc}
\hline
Dataset     & Attribute & Object & Composition &Images   & Avg. attr       & Avg. obj &Avg. resolution & Multi-attribute   \\ \hline\hline
UT-Zappos~\cite{yu2014fine}   & 16     & 12     & 149 &  29,126       & 24.75     & 24.75     & 136$\times$102       & \cxmark      \\
MIT-States~\cite{isola2015discovering} & 115   & 245  & 1,962 & 53,753  &  8.01    & 17.06   & 256$\times$256 & \cxmark            \\
C-GQA~\cite{naeem2021learning}      & 413    & 674   & 7,767 &39,298    &  18.81   & 11.52    & 286$\times$223 & \cxmark            \\
\rowcolor[gray]{0.975} MAC (Ours)       & 99        & 259    & \textbf{17,627}    &22,838     & \textbf{31.41}   & \textbf{82.18}              & \textbf{1092$\times$982} & \checkmark       \\\hline
\end{tabular}
\label{tab:dataset}
\vspace{-5mm}
\end{table*}

%% file: tabs/table_main_result.tex
\begin{table*}[!t]
\centering
\caption{\textbf{Main results on MAC.} CGE~\cite{naeem2021learning} uses ResNet18 pretrained on ImageNet, while other methods use CLIP ViT-L/14. 
\dag ~ indicates that the results are obtained without training on MAC.
}
\resizebox{\textwidth}{!}{

\begin{tabular}{l|cccccc|ccc}
\hline
            Method & Exact Match  & Top1-P  & Top5-R  & Coverage $\downarrow$ & Top1-P-attr & Top1-P-obj  &AUC &Seen & Unseen\\ \hline\hline  
\multicolumn{10}{c}{  \it{\textbf{Closed-world Results} }} \\ \hline
CGE~\cite{naeem2021learning}     & 4.90  & 29.85 & 33.32 & 155.63 & 47.19 & 59.75  &5.59 &32.50 &24.45  \\
CLIP\dag~\cite{radford2021learning}  & 5.92  & 42.42 & 39.98 & 69.14 & 51.03 & 80.23 &18.81&43.96&51.95   \\
CoOP~\cite{zhou2022learning}   & 6.51  & 45.81 & 39.86 & 46.08 & 54.45 & 80.82  &18.65&45.47&51.09    \\
CSP~\cite{nayak2022learning}      & 5.38  & 39.01 & 37.26 & 36.36 & 48.26 & 77.52 &15.78&40.43&48.18  \\
DFSP~\cite{lu2023decomposed}       & 7.82  & 46.76 & 50.35 & \textbf{25.08} & 53.03 & 85.74& 20.47&48.33&53.89  \\
GIPCOL~\cite{xu2024gipcol}       & 6.01  &41.62   &37.40 &29.55  &48.45 &84.22&14.22&42.87&42.58 \\
Troika~\cite{huang2023troika}     & 12.79  & 62.58 & 56.73 & 26.00 & 69.85 & \textbf{87.77} &25.93& 64.89&  \textbf{54.87}  \\ 

LLAVA-7B\dag~\cite{liu2024llava}     & 7.69 & 43.50  & 35.10   & 282.47 & 63.02  & 66.89&10.35&43.01&33.82 \\
LLAVA-13B\dag~\cite{liu2024llava}     &7.38 & 41.71  &35.66  &248.65 & 61.39  &69.52&10.34 &45.27&31.27  \\
\rowcolor[gray]{0.975} \methodname         & \textbf{17.42} & \textbf{63.45} &\textbf{62.37} & 31.35 & \textbf{71.55} &  86.45& \textbf{27.89}&\textbf{65.91}&50.85 \\
\hline

\multicolumn{10}{c}{  \it{\textbf{Open-world Results}} } \\ 
\hline

CGE~\cite{naeem2021learning}  & 4.37 & 27.86 & 30.26 & 448.29 & 44.79 & 59.52 &1.63&32.41&7.42   \\
CLIP\dag~\cite{radford2021learning}          & 3.19 & 29.37 & 27.88  & 229.49 & 37.37  & 78.93    &5.11&43.98& 17.03\\
CoOP~\cite{zhou2022learning}              & 4.60 & 36.11 & 28.47   & 144.21  & 44.94  & 79.81&5.91&45.41&17.52   \\
CSP~\cite{nayak2022learning}              & 2.77 & 26.20 & 24.87  & 360.35 & 34.16 & 76.30 &4.21&40.45&16.18   \\
DFSP~\cite{lu2023decomposed}   & 6.34 & 41.74 & 43.48 & 64.40 & 49.05 & 84.71&7.37&48.33&21.29  \\
GIPCOL~\cite{xu2024gipcol}  &3.78   &30.80   &22.29   &102.44   &36.81  &83.70 &3.11&41.20&10.46  \\
Troika~\cite{huang2023troika}       &  11.05 &  61.32 & 49.75 & 78.03 & 68.70  & 85.50 &8.56 &64.92 &18.00   \\ 
LLAVA-7B\dag~\cite{liu2024llava}   & 6.70 & 41.49  & 32.54   & 640.31 & 61.28  & 66.76 &4.79&42.96&15.82\\
LLAVA-13B\dag~\cite{liu2024llava}  &6.53  & 39.56  & 33.06   &593.53  &59.45   &69.54&5.61&45.25&16.42  \\
\rowcolor[gray]{0.975} \methodname        &  \textbf{16.77} & \textbf{62.60}    & \textbf{61.00}   & \textbf{57.55} & \textbf{70.92}   & \textbf{86.32} &\textbf{13.02} &\textbf{65.89} & \textbf{25.18}  \\\hline
\end{tabular}
}

\label{tab:main_result}
\vspace{-5mm}
\end{table*}

%% file: tabs/tab_efficiency.tex
\begin{table}[t]  
\centering
\caption{\textbf{Efficiency analysis on open-world setting.} FLOPs quantify the computational load for a model on a single image. GPU memory is measured with a batch size of 16. Inference time is averaged over 1k randomly selected MAC samples.}  
\setlength\tabcolsep{10pt}
\begin{tabular}{l|rrr}   
\hline
Model & FLOPs  & GPU Memory  & Inference Time \\ \hline\hline
DFSP~\cite{lu2023decomposed}        &    19.0T      &    14995MB  &    746ms       \\  
Troika~\cite{huang2023troika}      &    18.3T      &    18973MB  &    1462ms        \\  
\rowcolor[gray]{0.975} \methodname  &    0.6T       &    2283MB   &    28ms     \\  
\hline  
\end{tabular}  
\label{tab:efficiency}  
\vspace{-5mm}
\end{table}  

%% file: tabs/tab_cross.tex
\begin{table}[t]
\setlength\tabcolsep{5.1pt}
\caption{\textbf{Cross-Dataset Evaluation Results.} 
Models trained on MAC are directly evaluated  on the test set. }
\begin{tabular}{l|c|c|cc}
\hline
    \multirow{2}{*}{Method} & \multirow{2}{*}{Training Set} & \multirow{2}{*}{Testing Set}  & \multicolumn{2}{c}{Metric}  \\ \cline{4-5}
    & &  & Top-3-Pair & Top-1-Pair  \\ \hline\hline

     \multirow{2}{*}{Ours} & MIT-States  & \multirow{8}{*}{MIT-States} & 57.0  & 28.9  \\
     & MAC &   & 70.4\textcolor[gray]{0.4}{(+13.4)} & 34.9  \textcolor[gray]{0.4}{(+6.0)} \\ \cline{1-2} \cline{4-5}

    \multirow{2}{*}{Tri-CoOP} & MIT-States &  & 67.5  & 39.1  \\   
    & MAC&  & 70.4 \textcolor[gray]{0.4}{(+2.9)} & 39.3 \textcolor[gray]{0.4}{(+0.2)} \\  \cline{1-2} \cline{4-5}

   \multirow{2}{*}{CoOP~\cite{zhou2022learning}} & MIT-States & & 70.7  & 41.2   \\
    & MAC& & 68.9\textcolor[gray]{0.4}{(-1.8)} & 39.5\textcolor[gray]{0.4}{(-1.7)}  \\ \cline{1-2} \cline{4-5}

   \multirow{2}{*}{Troika~\cite{huang2023troika}} & MIT-States&  & 71.8  & 44.3  \\   
    & MAC&  & 70.0 \textcolor[gray]{0.4}{(-1.8)} & 37.4 \textcolor[gray]{0.4}{(-6.9)}  \\ \cline{1-2} \cline{4-5}
    
    \hline
\end{tabular}
\label{tab:cross}
\vspace{-5mm}
\end{table}

%% file: tabs/tab_ablation.tex


\begin{table}[t]
\tiny
\tablestyle{5pt}{1.0}
\setlength\tabcolsep{6pt}
\caption{\textbf{Ablation study on \methodname.}  Group 1 is the zero-shot result of CLIP. ``Single'' represents the composition  branch, and ``Double'' indicates the branches for attribute and object. ``CLS'' and ``Patch'' refer to the source of image token fed into the multimodal fusion module. Group 7 is the full version of \methodname.}
\label{tab:ablation}
\begin{tabular}{c|cc|cc|ccc}
\hline
\multirow{2}{*}{Group}  & \multicolumn{2}{c|}{Prompt Tuning} & \multicolumn{2}{c|}{Fusion } & \multicolumn{3}{c}{Metric} \\ \cline{2-8}
  & Single & Double & CLS & Patch & Exact & obj & attr \\ \hline\hline
1  &   &   &   &   & 5.92 & 80.23 & 51.03 \\ 
2  & \checkmark &   &   &   & 6.53 & 80.95 & 54.93 \\ 
3 &   & \checkmark &   &   & 9.20 & 85.63 & 60.42 \\ 
4 &   & \checkmark & \checkmark &    & 16.49  & 86.20 &  71.10\\ 
5  &   & \checkmark &   & \checkmark & 15.78 & 85.23 & 68.19 \\ 
6  &   &    & \checkmark & \checkmark & 16.64 & 86.09 &  70.61 \\ 
\rowcolor[gray]{0.975} 7 &   & \checkmark & \checkmark & \checkmark & \textbf{17.42} & \textbf{86.45} & \textbf{71.55} \\ \hline
\end{tabular}
\vspace{-5mm}
\end{table}

